\documentclass[11pt]{article}

\usepackage[final]{acl}

\usepackage{times}
\usepackage{latexsym}
\usepackage[T1]{fontenc}
\usepackage[utf8]{inputenc}
\usepackage{microtype}
\usepackage{inconsolata}
\usepackage{graphicx}
\usepackage{hyperref}
\usepackage{url}
\usepackage{subcaption}
\usepackage{caption}
\usepackage{float}
\usepackage{wrapfig}
\usepackage{xcolor}
\usepackage{colortbl}
\usepackage{booktabs}
\usepackage{multirow}
\usepackage{mdframed}
\usepackage{tabularx}
\usepackage{arydshln}
\usepackage{amsmath}
\usepackage{amssymb}
\usepackage[linesnumbered,ruled,vlined]{algorithm2e}

\title{Beyond Black-Box Interventions: Latent Probing for Faithful Retrieval-Augmented Generation}

\author{
    Linfeng Gao\textsuperscript{1,4}\thanks{\,\, Work done during internship at Alibaba Group.} \quad
    Qinggang Zhang\textsuperscript{2}\thanks{\,\, Corresponding author} \quad
    Baolong Bi\textsuperscript{3} \quad
    Bo Zeng\textsuperscript{4} \quad
    Zheng Yuan\textsuperscript{2} \quad
    \textbf{Zerui Chen}\textsuperscript{1} \quad \\
    \textbf{Zhimin Wei}\textsuperscript{1} \quad
    \textbf{Shenghua Liu}\textsuperscript{3} \quad
    \textbf{Linlong Xu}\textsuperscript{4} \quad
    \textbf{Longyue Wang}\textsuperscript{4} \quad
    \textbf{Weihua Luo}\textsuperscript{4} \quad
    \textbf{Jinsong Su}\textsuperscript{1,5}\footnotemark[2] \quad \\
    \textsuperscript{1}School of Informatics, Xiamen University
    \textsuperscript{2}The Hong Kong Polytechnic University \\
    \textsuperscript{3}University of Chinese Academy of Sciences
    \textsuperscript{4}Alibaba Group \\
    \textsuperscript{5}Key Laboratory of Digital Protection and Intelligent Processing of Intangible \\
Cultural Heritage of Fujian and Taiwan (Xiamen University), Ministry of Culture and Tourism, China\\
    \texttt{gaolinfeng@stu.xmu.edu.cn, qinggang.zhang@connect.polyu.hk, 
    jssu@xmu.edu.cn}
}

\begin{document}
\maketitle
\begin{abstract}
Retrieval-Augmented Generation (RAG) systems often fail to maintain contextual faithfulness, generating responses that conflict with the provided context or fail to fully leverage the provided evidence. Existing methods attempt to improve faithfulness through external interventions, such as specialized prompting, decoding-based calibration, or preference optimization. However, since these approaches treat the LLM as a black box, they lack a reliable mechanism to assess when and why knowledge conflicts occur. Consequently, they tend to be brittle, data-intensive, and agnostic to the model's internal reasoning process. In this paper, we move beyond black-box interventions to analyze the model's internal reasoning process.  We discover that conflicting and aligned knowledge states are linearly separable in the model’s latent space, and contextual noise systematically increases the entropy of these representations. Based on these findings, we propose ProbeRAG, a novel framework for faithful RAG that operates in three stages: (i) fine-grained knowledge pruning to filter irrelevant context, (ii) latent conflict probing to identify hard conflicts in the model's latent space, and (iii) conflict-aware attention to modulate attention heads toward faithful context integration. Extensive experiments demonstrate that ProbeRAG substantially improves both accuracy and contextual faithfulness. 
The related resources are available at \textcolor{blue}{\url{https://github.com/XMUDeepLIT/ProbeRAG}}.


\end{abstract}

\section{Introduction}

\begin{figure}
    \centering
    \includegraphics[width=1\linewidth]{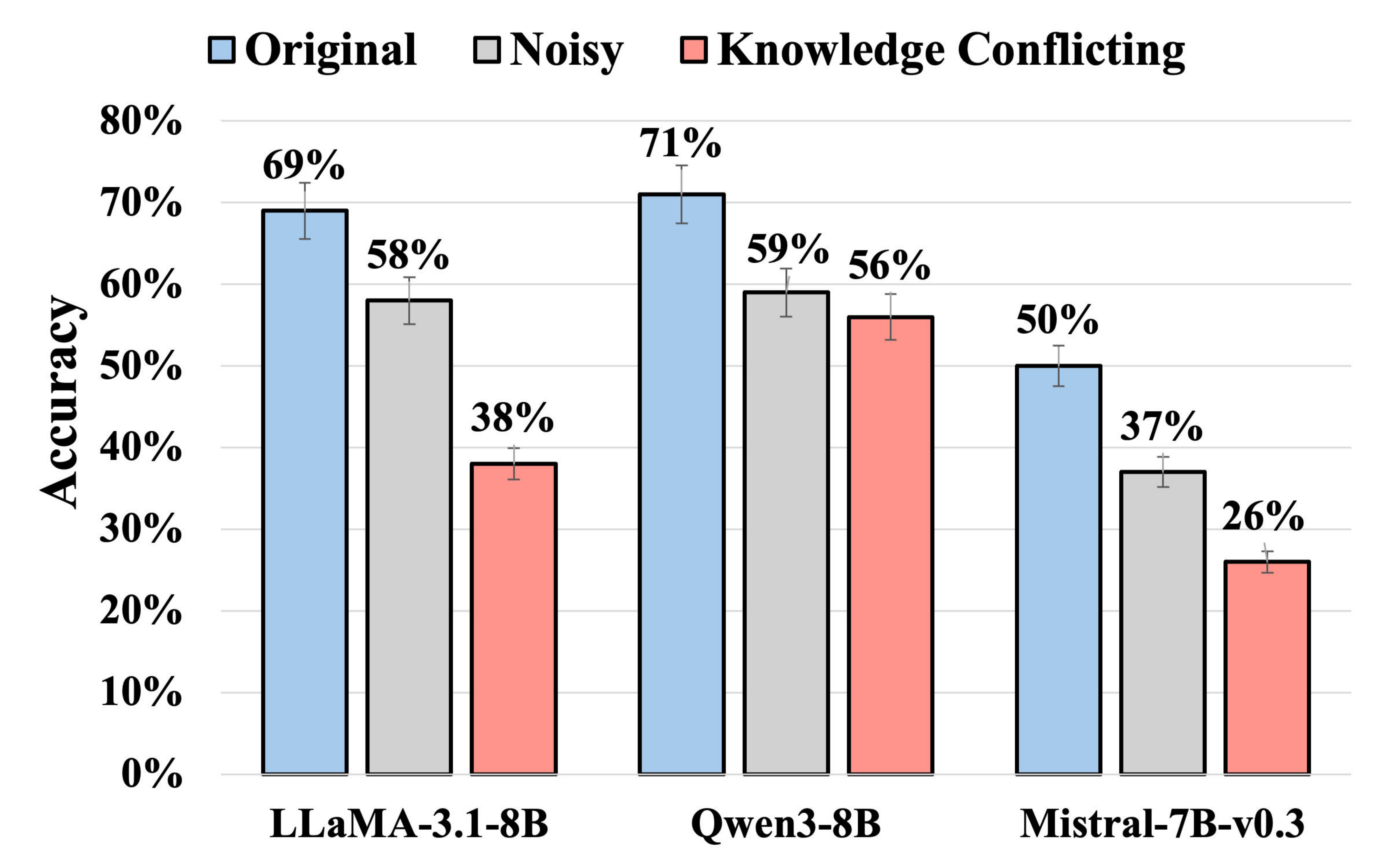}
    \caption{Experiments reveal that RAG systems degrade the performance when (i) exposed to contextual noise or (ii) confronted with conflicting knowledge.}
    \label{fig:pre_exp}
    \vspace{-4mm}
\end{figure}

Retrieval-Augmented Generation (RAG) has rapidly evolved as a powerful paradigm to enhance Large Language Models (LLMs) with external, up-to-date knowledge, effectively mitigating the problem of outdated or hallucinated knowledge~\citep{guu2020realm,feng2024retrieval,zhang2025survey}. Despite its promise, existing RAG systems often struggle with a critical challenge of contextual faithfulness in practice~\citep{bi2024context, bi2024factuality}, generating responses that are inconsistent with the retrieved context or fail to fully leverage the provided external knowledge~\citep{xu2024knowledge,zhang2025faithfulrag}. When faced with noisy or conflicting context, the LLM’s knowledge integration process fails, causing the model to default to its parametric knowledge or produce incoherent hybrids. As shown in Figure~\ref{fig:pre_exp}, even Qwen3-8B suffers a significant drop in accuracy, from 71\% to 59\% when the context contains noise, and further to 56\% when confronted with conflicting knowledge.

Existing methods to improve faithfulness primarily rely on external interventions, which can be classified into three categories. (i) Prompt-based methods, which design specialized instructions to guide the model’s reasoning process~\citep{zhou-etal-2023-context,selfrag,intuitive,zhang2025faithfulrag}. While this strategy can indeed improve factual grounding, its performance is often highly sensitive to the prompt and can not generalize across different domains or tasks. (ii) Decoding-based models that improve RAG faithfulness by adjusting the decoding process, calibrating the logits with contextual scoring~\citep{cad,coiecd}. However, these methods are often tightly coupled with specific decoding strategies and become brittle when the retrieved context contains noise or contradictions. (iii) Preference optimization methods fine-tune the LLM using Direct Preference Optimization (DPO) to encourage faithful grounding~\citep{si2025teaching,bi2024context}. Although enabling end-to-end learning, these methods depend heavily on carefully designed reward functions and large-scale, high-quality preference data, which are costly to collect.

More crucially, existing methods share a fundamental limitation: they treat the LLM as a black box, applying external interventions without understanding the internal mechanisms during the reasoning process. Consequently, their interventions remain correlational rather than causal: they may statistically associate certain inputs with more faithful outputs, but cannot diagnose why the model fails in specific conflict instances, nor predict its behavior under novel forms of contradiction. To achieve robust faithfulness, we argue it is necessary to move beyond external corrections and focus on two critical questions: how is the conflict represented within the model's latent space, and how does the latent conflict disrupt faithful generation? 

In this work, we conduct a latent space analysis to uncover how LLMs internally integrate contextual knowledge with their parametric memory and how they represent conflicting knowledge within their latent space. Specifically, we uncover two critical findings: (i) internal hidden-state of conflicting and aligned knowledge are linearly separable in the model’s latent space, providing a latent feature for conflict detection; and (ii) contextual noise systematically increases the entropy of these representations, which clarifies why noisy contexts always obscure the latent conflict feature.

Motivated by these findings, we introduce ProbeRAG, a framework that leverages latent probing to detect and mitigate conflicts for faithful RAG. It consists of three steps: (i) fine-grained knowledge pruning to reduce noise by filtering irrelevant context, (ii) latent conflict probing, where a lightweight probe is used to model the mapping relationship between hidden states and conflicting/aligned knowledge, and (iii) conflict-aware attention to modulate attention heads based on probe outputs to ensure faithful grounding. By integrating latent-space probing with targeted intervention, ProbeRAG shifts the paradigm from external constraint to internal guidance, enabling more robust, generalizable faithfulness. Our contributions are summarized as follows:

\begin{itemize}
    \item We perform an in-depth analysis on the models' internal knowledge integration mechanism. We discover the latent conflict feature existed in this process and the negative obscurity of contextual noise.
    \item We propose ProbeRAG, a framework designed to enhance RAG faithfulness, which detects latent conflicting knowledge and further guides the model to pay more attention while integrating the context.
    \item We evaluate the effectiveness of our framework on multiple benchmarks, demonstrating that ProbeRAG consistently outperforms competitive baselines.
\end{itemize}

\section{Preliminary Study}
\begin{figure}
\centering
\includegraphics[width=1\linewidth]{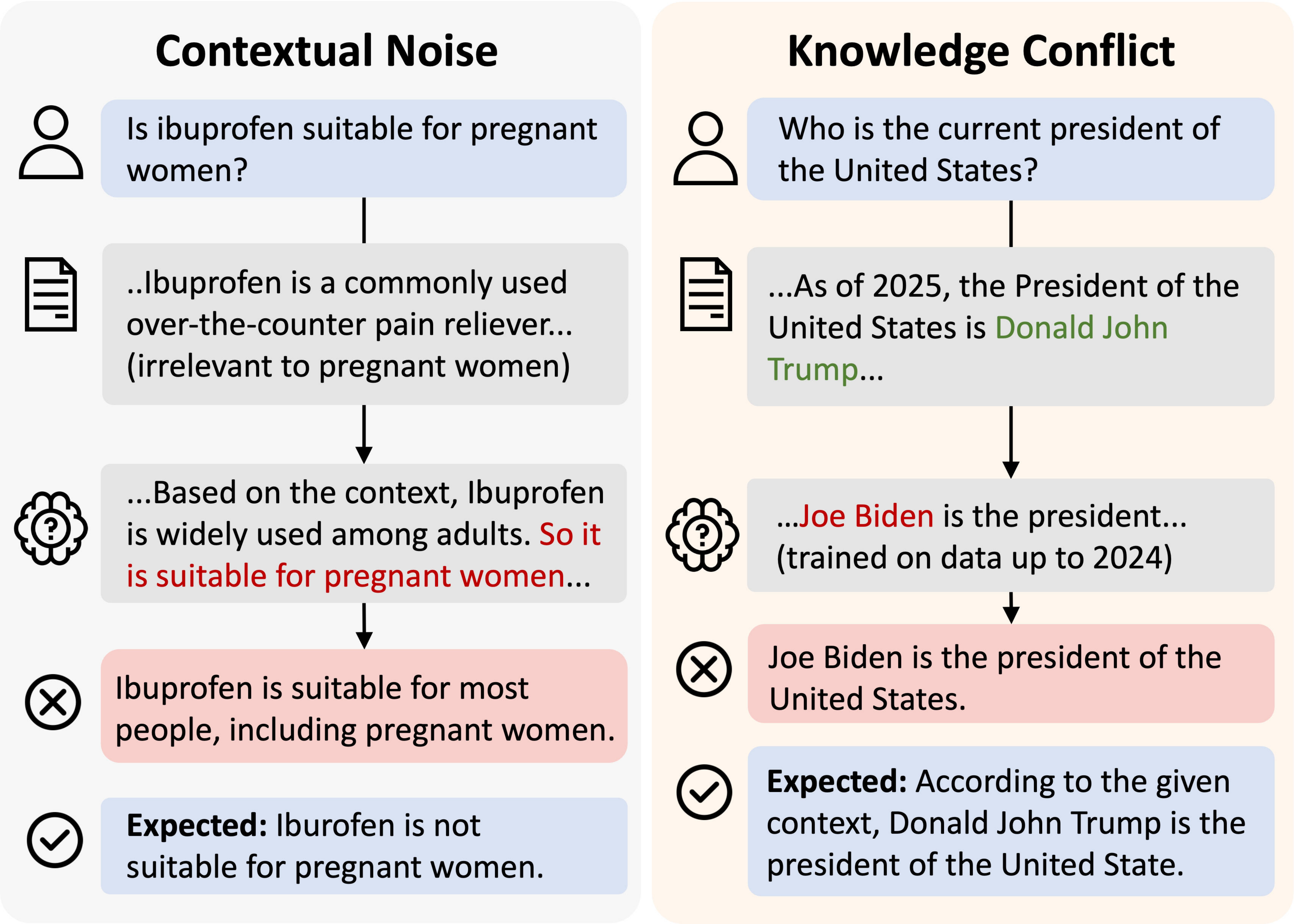}
\caption{Contextual noise draws the model's attention and enhances its inherent knowledge (in the left case). And the model prefers to follow inherent knowledge in the conflicting scenario (in the right case).}
\label{fig:pre_case}
\vspace{-3mm}
\end{figure}



\begin{figure*}[t]
    \centering
    \begin{minipage}{0.63\textwidth}
        \centering
        \includegraphics[width=\linewidth]{resources/preliminary/hidden_state.pdf}
        \caption{T-SNE visualization of hidden-state patterns between aligned and conflicting knowledge. There is a clear distinction in the distribution of hidden states between aligned and conflicting knowledge. Additionally, in the noisy scenario, the distinct pattern becomes unclear, emphasizing the importance of noise filtering.}
        \label{fig:hidden}
    \end{minipage}
    \hfill 
    \begin{minipage}{0.32\textwidth}
        \centering
        \vspace{30pt} 
        \small 
        \begin{tabularx}{\linewidth}{lXXXX} 
            \toprule
            \textbf{Layer} & \textbf{0.6B} & \textbf{4B} & \textbf{8B} & \textbf{32B} \\ \midrule
            \textbf{L2}  & 0.213 & 0.252 & 0.324 & 0.267 \\
            \textbf{L4}  & 0.294 & 0.315 & 0.545 & 0.318 \\
            \textbf{L8}  & 0.221 & 0.324 & 0.232 & 0.458 \\
            \textbf{L16} & \textbf{0.343} & 0.246 & 0.474 & 0.663 \\
            \textbf{L32} & N/A   & \textbf{0.432} & \textbf{0.655} & 0.480 \\
            \textbf{L64} & N/A   & N/A   & N/A   & \textbf{0.698} \\ \bottomrule
        \end{tabularx}
        \vspace{5mm}
        \captionof{table}{The JS Divergence of the hidden-state distribution between aligned knowledge and conflict knowledge. We compare different layers and different model size.}
        \label{tab:js-divergence}
    \end{minipage}
    \vspace{-2mm}
\end{figure*}

\label{sec:preliminary}

\subsection{Existing Challenges in RAG Faithfulness}

Based on the consensus reached through existing works~\cite{ji2023survey, bi2024context, zhang2025faithfulrag, yuan2025exploiting}, there are two key factors that underlie this issue: (i) Knowledge conflict between the contextual and internal knowledge of the model. Since the models tend to prioritize their parametric memory over the external evidence. (ii) Contextual noise with weak relevance to the task. As the model's attention will be drawn to this noise. To assess these two factors, we designed two controlled scenarios. In the \textit{knowledge conflicting} scenario, we select several key entities in the context and replace them with other entities of the same type to construct counterfactual knowledge. By doing these, we introduce inconsistencies between the context and model knowledge to construct knowledge conflicts. In the \textit{noisy} scenario, the original context is augmented with passages that are semantically aligned with the query but topically irrelevant, introducing unrelated knowledge. 

As shown in Figure~\ref{fig:pre_exp}, we conduct evaluations on three different open-source models, and all their accuracy decreases significantly in both scenarios. Some typical error cases are presented in Figure~\ref{fig:pre_case}. In the contextual noise scenario, their accuracy drops by more than 10\%. As demonstrated in the case studies, the primary issue is that irrelevant context amplifies non-task-related knowledge within the model, allowing it to gain undue priority during generation. In the knowledge conflict scenario, the model performance decrease is more pronounced, particularly for LLaMA-3.1-8B-Instruct and Mistral-7B-v0.3. Case study indicates that the model assigns a lower priority to contexts involving conflicts within its internal knowledge.

\subsection{Latent Space Probing and Analysis}

To further explore how models integrate external knowledge, we analyze the knowledge representations in their latent space. Following the method of~\citep{Xie2024KnowledgeConflict}, we extract the model’s parametric knowledge $K_a$, and construct conflicting knowledge $K_c$. We input each knowledge pair $\langle K_a, K_c \rangle$ into the models and extract their hidden states to perform a two-dimensional visualization using t-SNE~\citep{van2008visualizing}. The internal representations are visualized in Figure~\ref{fig:hidden} (\textcolor[rgb]{1, 0.5, 0.5}{red} and \textcolor[rgb]{0.5, 0.6, 0.9}{blue} points). From the figure, we observe that aligned knowledge and conflicting knowledge form two clearly separable distributions, while noise will to some extent affect this separability. This observation indicates that a distinct ``conflict feature'' exists within the model’s latent space, manifesting as two linearly separable clusters of hidden states for aligned and conflicting knowledge. Grounded in the Linear Representation Hypothesis~\cite{park2023linear}, this phenomenon occurs because pre-trained models map knowledge consistent with their parameters into dense semantic clusters, whereas conflicting information induces a divergence in semantic vector directions~\cite{zhao2024analysing}. Due to the large angular differences between these distinct sources of information, conflicting states align more closely with a structural ``conflict feature'' feature than with specific semantic content~\cite{park2023linear}. This geometric separation explains how internal biases are generated during knowledge integration, leading the model to favor consistent knowledge while overlooking conflicts. By extending these theoretical insights, our framework aims to unlock the model’s ``black box'' and enhance its overall faithfulness.

\begin{figure*}
    \centering
    \includegraphics[width=\linewidth]{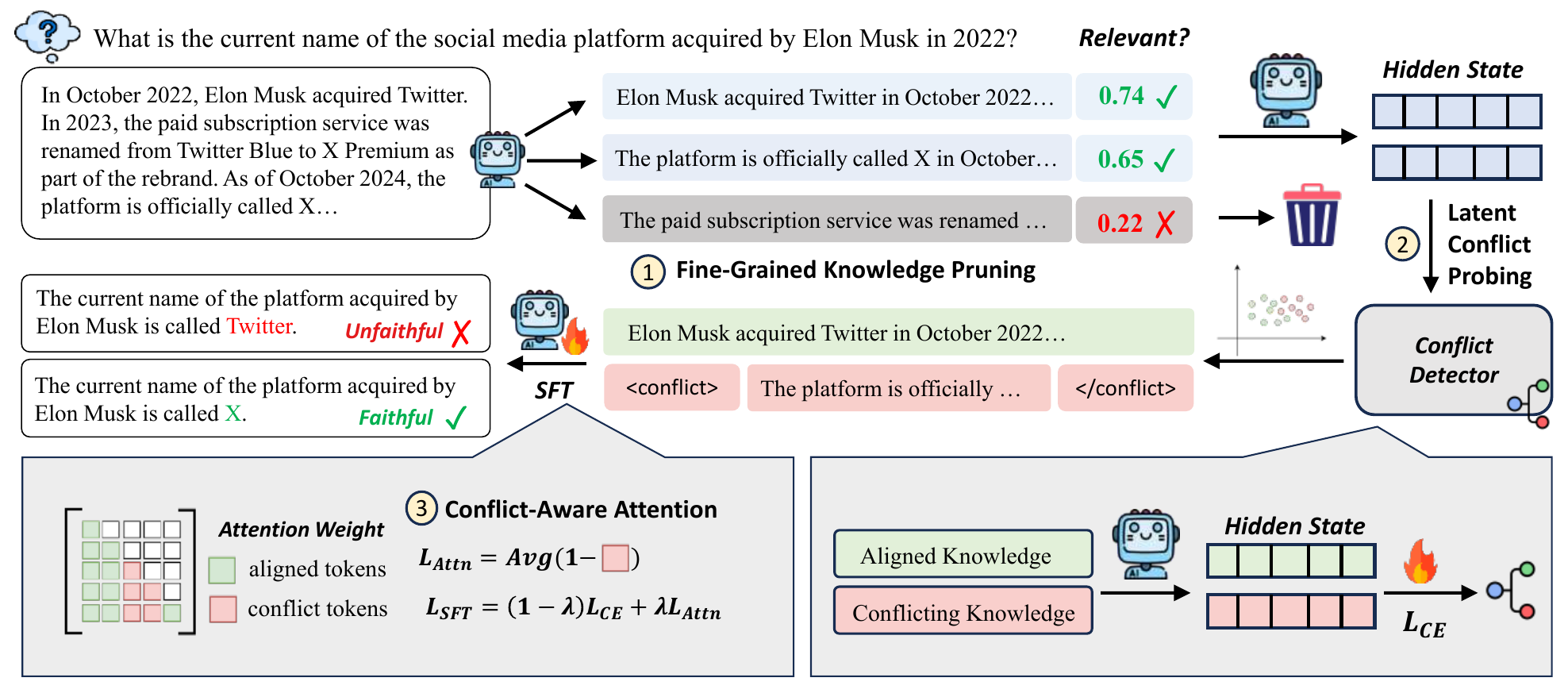}
    \caption{The overview of the framework ProbeRAG, which comprises three steps: (1) \textbf{Fine-Grained Knowledge Pruning}: decompose the context and filters out irrelevant knowledge; (2) \textbf{Latent Conflict Probing}: detect knowledge conflict via a probe trained in the model's latent space; (3) \textbf{Conflict-Aware Attention}: regulate the model's attention distribution on conflicting knowledge via fine-tuning.}
    \label{framework}
    \vspace{-4mm}
\end{figure*}

Additionally, we analyzed the distribution of the hidden state JSD across different layers in Qwen models of varying sizes. As shown Table~\ref{tab:js-divergence}, there is a clear trend of increasing JSD as information propagates through deeper layers. This reflects the hierarchical nature of transformer representations~\cite{tenney2019bert, belinkov2022probing}: lower layers typically encode surface-level features, while middle-to-late layers capture higher-level semantic abstractions and long-range dependencies. Since knowledge conflicts are often more abstract semantic features, they are typically captured only at higher levels of the model. Moreover, larger models exhibit more pronounced ``conflict features'', consistent with the scaling law of models~\cite{kaplan2020scaling}: larger models possess stronger semantic understanding capabilities, thereby more readily capturing ``conflict features''.

\section{Method}
\label{sec:method}
In this section, we introduce the framework ProbeRAG, which focus on the internal guidance instead of external interventions. As illustrated in Figure~\ref{framework}, it comprises three steps: (1) \textbf{Fine-Grained Knowledge Pruning}: where the context is decomposed into fine-grained knowledge and irrelevant knowledge is filtered out; (2) \textbf{Latent Conflict Probing}: we detect knowledge conflict via a probe trained in the model's latent space. It takes the model's hidden state as input and predicts whether the contextual knowledge is in conflict with the model's parametric knowledge; (3) \textbf{Conflict-Aware Attention}: we fine-tune the model to regulate its attention head towards faithful context integration via an auxiliary attention-guidance loss item. the following subsections will provide detailed illustrations for each step.

\subsection{Fine-Grained Knowledge Pruning}
Considering the interference factors of noise on latent space ``conflict features'', the most appropriate unit for processing is the fine-grained knowledge, which corresponds to an independent, complete sentence-level statement that cannot be further decomposed. Each statement preserves the subject–predicate–object structure with necessary modifiers, ensuring no information is lost during decomposition. To extract knowledge $\{K_1, K_2, \ldots,K_n\}$ from a given context $D$, we leverage the decomposition capabilities of an external LLM (we choose GPT-4o~\citep{openai_gpt4o_card} for its strong reasoning and text-processing abilities). We also provide an ablation study (Section~\ref{sec:ablation}) to utilize an open-source model (LLaMA3-8B-Instruct) in this step. The decomposition result of the context is a list of knowledge. Formally, we define this process as:
$$ 
\text{Decompose}(D) = \{K_1, K_2, \ldots,K_n\}
$$
where $K_i$ denotes the $i$-th knowledge statement. Detailed prompt is provided in Appendix~\ref{appendix:details}.

We further filter irrelevant knowledge to reduce contextual noise. For each knowledge statement $K_i$, we compute its similarity with the query $Q$:
$$ 
f(Q, K_i)=\langle q,k_i \rangle 
$$
where $q = Enc(Q)$ and $k_i = Enc(K_i)$ are vector embeddings of the query and the knowledge item, respectively, and $\langle \cdot, \cdot \rangle$ denotes cosine similarity. We employ the all-MiniLM-L6-v2\footnote{https://huggingface.co/sentence-transformers/all-MiniLM-L6-v2} for embedding. Finally, the top-$k$ results are selected.

\subsection{Latent Conflict Probing}
To effectively detect knowledge conflict, we utilize the latent conflict feature in the model as we observed in our preliminary study in Section~\ref{sec:preliminary}. Specifically, when aligned and conflicting knowledge are input into the model, the model's hidden state projects them into two separable clusters. Leveraging this property, we implement a reverse prediction through a probe: by inputting the model's hidden state, it determines whether the knowledge is aligned or conflicting. To enable the probe to fit the reverse process of the model's hidden state mapping, the probe model must be trained in the model's latent space. We leverage the knowledge editing dataset MQuAKE~\citep{zhong2023mquake}, since it contains both aligned knowledge prior to editing and conflicting knowledge after editing, providing natural pairs of aligned and conflicting knowledge $\langle K_a, K_c \rangle$. Importantly, the data format and textual granularity in MQuAKE align closely with the knowledge statements extracted in our framework, making it suitable for supervision. 

During inference, the resulting filtered knowledge from previous step is passed through the frozen model to obtain its hidden state representation, which is subsequently classified by the probe:
$$
\mathcal{M}(K_i)\in \mathbb{R}^{d_M}, \quad \mathcal{P}\big(\mathcal{M}(K_i)\big)\in\{0,1\},
$$
where $\mathcal{M}(K_i)$ denotes the hidden state of knowledge statement $K_i$ produced by frozen model $\mathcal{M}$ with dimension $d_M$ and $\mathcal{P}$ denotes the probe. 

We validated the accuracy and generalization of using probe models for conflict detection in Section~\ref{sec:probe_eval}. Despite being trained on knowledge editing datasets, the probes maintained strong generalization capabilities on RAG domain data, ensuring the reliability of conflict detection. To further enhance the priority of conflicting knowledge in the model generation process, we mark the conflicting knowledge with special tokens, i.e., wrapping them within $\langle conflict \rangle$ and $\langle /conflict \rangle$. This explicit annotation enables the subsequent stage to be aware of which knowledge statement are conflicting and should be augmented. 

\subsection{Conflict-Aware Attention}
In this step, we aim to encourage the model to pay more attention to conflicting knowledge. External interventions like in-context learning can hardly replace the model's inherent knowledge, which is demonstrated in our ablation study in Section~\ref{sec:ablation}. Therefore, we attempt to leverage the attention mechanism within the model to enhance the contextual faithfulness and propose Conflict-Aware Attention. We introduce an additional attention-guidance loss term that explicitly regularizes the model’s attention distribution. Specifically, for each conflicting knowledge item $K_i$, we denote its token sequence as $T^{(i)}=\{t^{(i)}_1, t^{(i)}_2,\ldots,t^{(i)}_m\}$. The positions of these tokens in the input context are represented by $S=\{j \mid \exists \mathcal{P}(\mathcal{M}(K_i))=1, x_j \in T^{(i)}\}$, where $\mathcal{P}(\mathcal{M}(K_i))=1$ indicates that knowledge item $K_i$ is judged as conflicting by the probe, and $x_j$ denotes the $j$-th token of the context. In practice, these positions in $S$ can be directly identified via the previously introduced special tokens $\langle conflict \rangle$ and $\langle /conflict \rangle$.

Based on this alignment, we extract the attention weights from subsequent tokens attending to the conflict-related tokens and then compute the attention guidance loss as follows:
\begin{align*}
    \mathcal{L}_{\text{Attn}} &= \frac{1}{|P|}\sum_{(i,j)\in P}(1-\alpha_{ij}), (i,j)\in P \\
    P&=\{(i,j)\mid i \geq j;\, j\in S\}
\end{align*}
where $\alpha_{ij}$ denotes the attention weight of token $i$ on token $j$. Finally, we combine the attention loss with the standard language modeling objective through a weighted sum:
$$
\mathcal{L}_{\text{SFT}} = (1-\lambda)\mathcal{L}_{\text{CE}} + \lambda \mathcal{L}_{\text{Attn}},
$$
where $\lambda \in [0,1]$ balances the trade-off to avoid overfitting to the conflicting knowledge. We provide an analysis on its impact in Section~\ref{sec:attn_analysis}. This joint objective ensures that the model not only learns to generate faithful outputs but also explicitly attends to conflicting knowledge during training.

\section{Experiment}
\begin{table*}[t]
\centering
\small
\vspace{-2mm}
\resizebox{\textwidth}{!}{
\begin{tabular}{llcccccccccc}
\toprule
\multirow{2}{*}{Category} & \multirow{2}{*}{Method} & \multicolumn{2}{c}{FaithEval} & \multicolumn{2}{c}{ConFiQA (MC)} & \multicolumn{2}{c}{ConFiQA (MR)} & \multicolumn{2}{c}{ConFiQA (QA)} & \multicolumn{2}{c}{SQuAD} \\
\cmidrule(lr){3-4} \cmidrule(lr){5-6} \cmidrule(lr){7-8} \cmidrule(lr){9-10} \cmidrule(lr){11-12}
 &  & F1 & EM & F1 & EM & F1 & EM & F1 & EM & F1 & EM \\
\midrule
\multicolumn{12}{c}{\cellcolor{blue!10}\textbf{LLaMA-3.1-8B-Instruct}} \\
\multirow{2}{*}{Baseline} & No-Context & 27.7 & 6.0 & 5.0 & 2.1 & 6.1 & 1.9 & 6.1 & 1.3 & 8.9 & 1.2 \\
 & Full-Context & 66.9 & 53.1 & 28.0 & 22.5 & 50.3 & 41.3 & 58.5 & 49.0 & 64.5 & 46.0 \\
 \midrule
\multirow{3}{*}{Prompt-Based}   & Opin(Instr)~\citep{zhou-etal-2023-context} & 34.9 & 15.1 & 67.4 & 57.3 & 65.9 & 54.0 & 76.9 & 67.4 & 66.0 & 47.7 \\
& KRE~\citep{ying2023intuitive} & 59.1 & 12.1 & 68.2 & 59.8 & 68.7 & 58.9 & 84.0 & 74.7 & 59.8 & 43.7 \\
& FaithfulRAG~\citep{zhang2025faithfulrag} & 62.4 & 49.1 & 70.3 & 61.2 & 62.4 & 53.1 & 70.4 & 65.2 & 61.3 & 50.2\\
\midrule
\multirow{3}{*}{Decoding-Based} & CAD~\citep{cad} & 59.4 & 42.7 & 16.0 & 11.4 & 40.0 & 31.3 & 48.3 & 38.1 & 60.3 & 41.8 \\
 & COIECD~\citep{coiecd} & 56.1 & 41.3 & 28.5 & 24.0 & 50.9 & 43.3 & 67.1 & 60.1 & 67.0 & 50.3 \\
 & AdaCAD~\citep{wang2025adacad} & 61.4 & 47.7 & 43.3 & 32.5 & 46.7 & 42.5 & 55.7 & 48.1 & 66.4 & 51.2 \\
 \midrule
\multirow{4}{*}{Training-Based} & Context-DPO~\citep{bi2024context} & 67.2 & 53.7 & 76.9 & 67.7 & 78.5 & 66.9 & 83.7 & 76.7 & 64.4 & 45.8 \\
 & CANOE~\citep{si2025teaching} & 71.6 & 56.3 & 80.9 & 74.2 & 80.2 & 72.6 & 82.3 & 77.7 & 65.4 & 49.7 \\
 & ParamMute~\citep{huang2025parammute} & 68.5 & 56.2 & 70.6 & 60.9 & 73.2 & 61.2 & 78.3 & 73.2 & 64.2 & 48.5 \\
  & ProbeRAG (ours) & \textbf{74.4} & \textbf{64.4} & \textbf{89.2} & \textbf{87.7} & \textbf{89.7} & \textbf{87.0} & \textbf{93.1} & \textbf{91.7} & \textbf{68.4} & \textbf{53.3} \\
\midrule
\multicolumn{12}{c}{\cellcolor{blue!10}\textbf{Qwen3-8B}} \\
\multirow{2}{*}{Baseline} & No-Context & 22.8 & 4.1 & 7.6 & 3.6 & 8.0 & 2.8 & 7.8 & 1.4 & 6.7 & 0.4 \\
 & Full-Context & 55.5 & 23.8 & 59.6 & 50.2 & 66.1 & 55.1 & 72.5 & 64.2 & 63.8 & 44.9 \\
 \midrule
\multirow{3}{*}{Prompt-Based} & Opin(Instr)~\citep{zhou-etal-2023-context} & 35.0 & 13.9 & 70.7 & 61.1 & 69.7 & 59.5 & 78.8 & 69.2 & 63.8 & 46.1 \\
   & KRE~\citep{ying2023intuitive} & 58.1 & 12.3 & 67.5 & 59.1 & 68.4 & 59.0 & 80.4 & 67.3 & 48.6 & 29.7 \\
   & FaithfulRAG~\citep{zhang2025faithfulrag} & 70.2 & 58.3 & 72.3 & 60.1 & 62.3 & 54.2 & 75.3 & 65.3 & 66.2 & 50.2 \\
\midrule
\multirow{3}{*}{Decoding-Based} & CAD~\citep{cad} & 57.0 & 28.7 & 57.7 & 48.3 & 64.8 & 53.3 & 71.0 & 62.0 & 63.6 & 44.5 \\
 & COIECD~\citep{coiecd} & 66.6 & 56.4 & 66.7 & 60.8 & 71.5 & 63.8 & 78.5 & 73.6 & 69.7 & 55.2 \\
 & AdaCAD~\citep{wang2025adacad} & 67.3 & 58.2 & 64.2 & 58.7 & 72.5 & 64.3 & 76.9 & 68.3 & \textbf{72.6} & \textbf{67.4} \\
 \midrule
\multirow{4}{*}{Training-Based} & Context-DPO~\citep{bi2024context} & 55.2 & 24.0 & 59.6 & 50.1 & 65.9 & 55.0 & 72.3 & 63.9 & 63.8 & 44.9 \\
 & CANOE~\citep{si2025teaching} & 70.3 & 60.2 & 85.2 & 81.7 & 84.6 & 80.7 & 92.2 & 86.5 & 69.4 & 53.4 \\
 & ParamMute~\citep{huang2025parammute} & 67.8 & 58.2 & 86.3 & 80.2 & 83.2 & 79.4 & 90.5 & 83.7 & 69.2 & 52.1 \\
 & ProbeRAG (ours) & \textbf{74.9} & \textbf{61.6} & \textbf{90.7} & \textbf{89.7} & \textbf{91.3} & \textbf{89.0} & \textbf{95.7} & \textbf{94.3} & 71.5 & 55.7 \\
\midrule
\multicolumn{12}{c}{\cellcolor{blue!10}\textbf{Mistral-7B-v0.3}} \\
\multirow{2}{*}{Baseline} & No-Context & 26.2 & 4.4 & 4.4 & 0.9 & 4.9 & 0.5 & 6.1 & 1.0 & 8.1 & 1.0 \\
 & Full-Context & 68.8 & 37.7 & 25.6 & 12.5 & 37.8 & 21.5 & 58.5 & 44.0 & 56.4 & 37.5 \\
 \midrule
\multirow{3}{*}{Prompt-Based} & Opin(Instr)~\citep{zhou-etal-2023-context} & 35.7 & 14.1 & 58.8 & 44.1 & 57.8 & 52.5 & 76.4 & 65.5 & 58.1 & 37.4 \\
   & KRE~\citep{ying2023intuitive} & 64.8 & 36.5 & 58.7 & 45.0 & 60.9 & 45.3 & 84.5 & 72.8 & 52.6 & 33.9 \\
   & FaithfulRAG~\citep{zhang2025faithfulrag} & 66.4 & 40.3 & 60.3 & 44.3 & 56.5 & 42.1 & 74.2 & 63.5 & 60.3 & 40.1\\
\midrule
\multirow{3}{*}{Decoding-Based} & CAD~\citep{cad} & 68.9 & 33.3 & 16.7 & 5.9 & 27.5 & 12.8 & 53.5 & 36.9 & 51.4 & 32.1 \\
 & COIECD~\citep{coiecd} & 64.4 & 29.5 & 26.1 & 14.5 & 39.3 & 26.3 & 58.9 & 45.1 & 59.2 & 39.7 \\
 & AdaCAD~\citep{wang2025adacad} & 68.4 & 34.2 & 33.7 & 23.4 & 39.6 & 29.4 & 58.2 & 46,3 & 61.6 & 43.6 \\
\midrule
\multirow{4}{*}{Training-Based} & Context-DPO~\citep{bi2024context} & 64.9 & 31.8 & 44.8 & 28.3 & 50.9 & 31.9 & 66.4 & 52.7 & 56.6 & 37.6 \\
 & CANOE~\citep{si2025teaching} & 64.1 & 44.9 & 87.2 & 85.7 & 84.7 & 81.9 & 92.5 & 90.7 & 57.8 & 42.5 \\
 & ParamMute~\citep{huang2025parammute} & 67.1 & 45.2 & 86.3 & 82.1 & 86.4 & 82.1 & 93.1 & 91.2 & 60.2 & 43.8\\
  & ProbeRAG (ours) & \textbf{74.9} & \textbf{62.9} & \textbf{91.2} & \textbf{89.7} & \textbf{90.8} & \textbf{88.2} & \textbf{95.1} & \textbf{93.7} & \textbf{68.1} & \textbf{53.6} \\
\bottomrule
\end{tabular}
}
\caption{Main results with methods grouped by Baseline, Prompt-Based, Decoding-Based, and Training-Based.}
\label{main_results}
\vspace{-4mm}
\end{table*}

\subsection{Setup}
\label{sec:setup}

\paragraph{Datasets.}

We evaluate ProbeRAG on three datasets. ConFiQA~\citep{bi2024context} is a benchmark designed to assess contextual faithfulness in question answering, particularly under real-world RAG scenarios involving knowledge conflicts. It consists of three subsets: QA (Question Answering), MR (Multi-hop Reasoning), and MC (Multi-Conflicts). QA is a single-hop question answering task, while MR and MC are multi-hop reasoning tasks in which the context includes one and multiple counterfactuals. FaithEval~\citep{ming2024faitheval} introduces conflicts at the level of logical reasoning: inconsistencies arise not from direct factual contradictions, but from reasoning chains that lead to conflicting conclusions. Finally, we also evaluate on SQuAD~\citep{rajpurkar2016squad}, following the version curated in KRE~\citep{ying2023intuitive}, which also incorporates fact-level knowledge conflicts.

\paragraph{Models and Baselines.}
We adopt several mainstream open-source models, including Llama-3.1-8B-Instruct, Qwen3-8B, and Mistral-7B-v0.3. We compare ProbeRAG against representative baseline methods from three major categories in the field of contextual faithfulness: prompt-based, decoding-based, and training-based approaches. Among the prompt-based methods, we include Opin(Instr)~\citep{zhou-etal-2023-context}, KRE~\citep{ying2023intuitive}, and FaithfulRAG~\citep{zhang2025faithfulrag}. For decoding-based methods, we evaluate CAD~\citep{cad}, COIECD~\citep{coiecd} and AdaCAD~\citep{wang2025adacad}. For training-based methods, we compare against Context-DPO~\citep{bi2024context}, CANOE~\citep{si2025teaching} and ParamMute~\citep{huang2025parammute}. Specifically, we partition the ConFiQA dataset into training and test sets. All baselines that require training are trained on the ConFiQA training set, and evaluation is consistently performed on the test set. Additional implementation details are provided in the Appendix~\ref{appendix:details}.

\begin{table*}[t]
  \centering
  \vspace{-2mm}
  \resizebox{\linewidth}{!}{
    \begin{tabular}{l l c c c c c c c c c c}
      \toprule
      \multirow{2}{*}{\textbf{Models}} & \multirow{2}{*}{\textbf{Modules}} & \multicolumn{2}{c}{\textbf{Faitheval}} & \multicolumn{2}{c}{\textbf{ConFiQA (MC)}} & \multicolumn{2}{c}{\textbf{ConFiQA (MR)}} & \multicolumn{2}{c}{\textbf{ConFiQA (QA)}} & \multicolumn{2}{c}{\textbf{SQuAD}}\\
      \cmidrule(lr){3-4} \cmidrule(lr){5-6} \cmidrule(lr){7-8} \cmidrule(lr){9-10} \cmidrule(lr){11-12}
      & & \textbf{F1} & \textbf{EM} & \textbf{F1} & \textbf{EM} & \textbf{F1} & \textbf{EM} & \textbf{F1} & \textbf{EM} & \textbf{F1} & \textbf{EM}\\
      \midrule
      \multirow{4}{*}{LLaMA-3.1-8B-Instruct} 
      & ProbeRAG & \textbf{74.4} & \textbf{64.4} & \textbf{89.2} & \textbf{87.7} & 89.7 & 87.0 & \textbf{93.1} & \textbf{91.7} & \textbf{68.4} & 53.3 \\
      & with Open Source KP & 70.3 & 59.5 & 83.2 & 82.4 & \textbf{90.3} & \textbf{87.2} & 90.4 & 89.8 & 65.7 & \textbf{54.2} \\
      & w/o Knowledge Pruning & 62.1 & 48.4 & 81.1 & 79.4 & 84.4 & 80.8 & 88.5 & 87.5 & 59.2 & 45.0 \\
      & w/o Latent Conflict Probing & 61.7 & 47.6 & 81.4 & 79.3 & 83.9 & 79.9 & 87.6 & 86.4 & 58.1 & 44.1 \\
      & w/o Conflict-Aware Attention & 61.5 & 50.9 & 83.8 & 80.4 & 85.0 & 81.0 & 87.5 & 86.4 & 58.2 & 40.2 \\
      \midrule
      \multirow{4}{*}{Qwen3-8B} 
      & ProbeRAG & \textbf{74.9} & \textbf{61.6} & \textbf{90.7} & \textbf{89.7} & \textbf{91.3} & \textbf{89.0} & \textbf{95.7} & \textbf{94.3} & \textbf{71.5} & \textbf{55.7} \\
      & with Open Source KP & 72.4 & 60.2 & 88.3 & 88.3 & 88.3 & 85.4 & 93.2 & 91.6 & 69.4 & 52.8 \\
      & w/o Knowledge Pruning & 62.6 & 50.9 & 86.1 & 85.3 & 86.7 & 85.2 & 88.8 & 87.8 & 66.3 & 51.3 \\
      & w/o Latent Conflict Probing & 61.0 & 49.8 & 85.4 & 84.6 & 86.6 & 85.1 & 88.6 & 87.5 & 66.1 & 51.0 \\
      & w/o Conflict-Aware Attention & 64.0 & 54.2 & 86.2 & 84.8 & 86.1 & 84.3 & 89.6 & 88.5 & 66.1 & 51.5 \\
      \midrule
      \multirow{4}{*}{Mistral-7B-v0.3} 
      & ProbeRAG & \textbf{74.9} & \textbf{62.9} & \textbf{91.2} & \textbf{89.7} & \textbf{90.8} & \textbf{88.2} & \textbf{95.1} & \textbf{93.7} & \textbf{68.1} & 53.6 \\
      & with Open Source KP & 70.3 & 60.7 & 88.5 & 85.7 & 87.3 & 84.9 & 92.5 & 90.4 & 66.7 & \textbf{54.3} \\
      & w/o Knowledge Pruning & 69.5 & 58.5 & 86.6 & 85.5 & 86.2 & 84.7 & 88.4 & 87.1 & 62.9 & 48.7 \\
      & w/o Latent Conflict Probing & 68.4 & 56.4 & 85.2 & 84.1 & 84.4 & 82.9 & 87.4 & 86.2 & 61.8 & 47.6 \\
      & w/o Conflict-Aware Attention & 69.3 & 57.6 & 88.8 & 86.1 & 86.3 & 81.8 & 81.4 & 77.4 & 59.7 & 49.8 \\
      \bottomrule
    \end{tabular}
  }
  \caption{Ablation Result. The ablation of each module significantly impacts the results. The Latent Conflict Probing module has the most substantial influence on the entire framework.}
  \vspace{-4mm}
  \label{tab:ablation}
\end{table*}

\subsection{Main Results}

As shown in Table~\ref{main_results}, ProbeRAG consistently achieves SOTA performance across all datasets and models. On FaithEval and ConFiQA (MC, MR, QA), ProbeRAG demonstrates strong generalization ability to both factual and logical conflicts, while on SQuAD, it further shows ProbeRAG improvements in traditional settings. Moreover, the consistent gains under different models highlight the robustness and generalizability of ProbeRAG.

Specifically, for model LLaMA-3.1-8B-Instruct, ProbeRAG achieves an F1 score of 74.4\% and an EM score of 64.4\% on FaithEval, outperforming the strongest baseline CANOE (71.6\% F1 / 56.3\% EM) by approximately +3\% F1 and +8\% EM. On ConFiQA sub-tasks, ProbeRAG improves over existing methods by 3\%–10\% across MC, MR, and QA, further confirming its robustness in handling conflict scenarios. Similarly, for Qwen3-8B, ProbeRAG attains 74.9\% F1 and 61.6\% EM on FaithEval, yielding substantial gains compared with prior methods, and reaches 90.7\% F1 and 89.7\% EM on the MC task. On Mistral-7B-v0.3, ProbeRAG achieves 74.9\% F1 / 62.9\% EM on FaithEval and great improvements across ConFiQA and SQuAD, surpassing the best training-based baselines by a clear margin. The consistent improvements across multiple datasets, conflict types, and backbone models underscore the effectiveness, robustness, and general applicability of ProbeRAG.

\subsection{Ablation Study}
\label{sec:ablation}
Ablation study results are summarized in Table~\ref{tab:ablation}. We first evaluate the effectiveness of performing Knowledge Pruning using an open-source LLM (with Open-Source KP). We substitute the GPT-4o with LLaMA-3-8B-Instruct, a lightweight, instruction-tuned open-source model. Results show that its ability to extract complete and semantically precise knowledge from the retrieved context is weaker than that of GPT-4o, which underscores the importance of context decomposition, as clean data facilitates subsequent process detection and enhancement. When the knowledge pruning step is removed, the model is forced to judge conflicts against every sentence in the context. Such coarse-grained filtering leads to incomplete contextual information and degrades the model’s ability to resolve fine-grained conflicts, thereby diminishing contextual faithfulness. More critically, removing the conflict detection module results in the most significant performance drop. Without explicit conflict detection, the downstream training becomes ineffective, since there are no targets to pay attention to. Finally, removing Conflict-Aware Attention also results in substantial degradation. Even when conflicts are annotated, the model struggles to prioritize them during inference due to its inherent tendency to rely on its parametric knowledge.

\begin{figure*}[t]
    \centering
    \includegraphics[width=\linewidth]{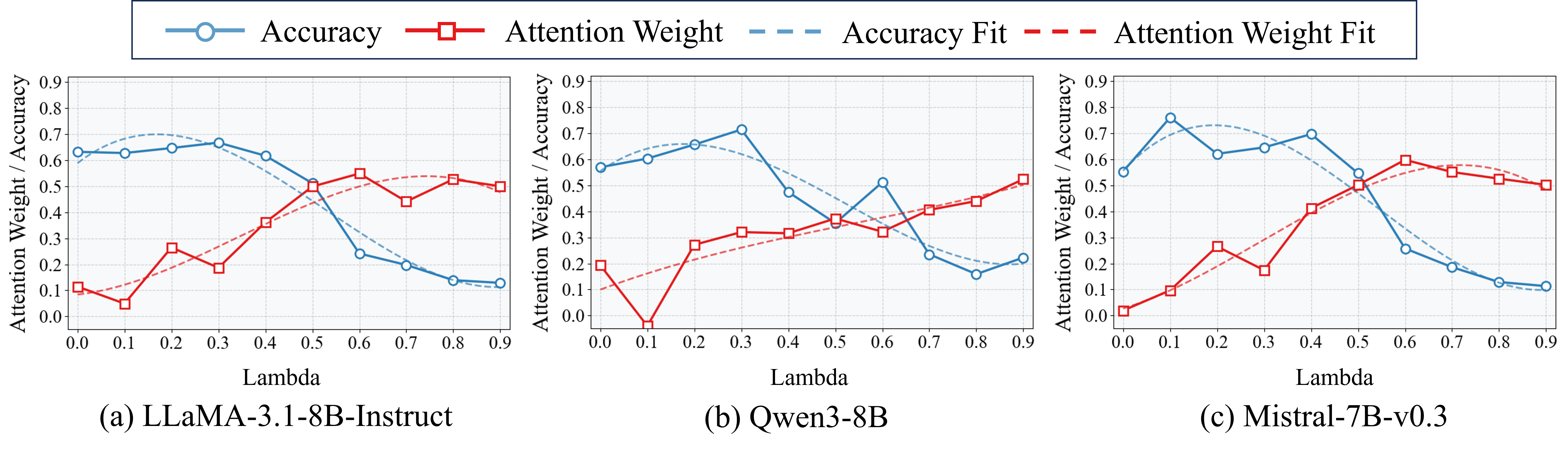}
    \caption{Impact of $\lambda$ on accuracy (\textcolor[rgb]{0.51,0.66,0.875}{blue}) and attention weight (\textcolor[rgb]{0.92, 0.45, 0.54}{red}). Performance peaks at smaller $\lambda$ values (0.1 to 0.3) and then declines, indicating that overfitting to conflicting knowledge degrades the performance.}
    \vspace{-4mm}
    \label{fig:atten}
\end{figure*}

\subsection{Evaluation on Conflict Detector}
\label{sec:probe_eval}

To further validate the probe model's generalization capabilities in conflict detection, we evaluated the probe model used in the main experiment across different benchmarks.

\begin{table}[ht]
\centering
\resizebox{\linewidth}{!}{
\begin{tabular}{lcccc}
\toprule
\textbf{Model} & \textbf{MQuAKE} & \textbf{FaithEval} & \textbf{ConFiQA} & \textbf{SQuAD} \\
\midrule
$\mathcal{P}_{Llama3.1}$ & 97.26 & 86.67 & 87.50 & 82.23 \\
$\mathcal{P}_{Qwen2.5}$ & 96.53 & 82.82 & 84.33 & 79.57 \\
$\mathcal{P}_{Mistral0.3}$ & 98.43 & 87.42 & 88.75 & 82.64 \\
\bottomrule
\end{tabular}
}
\caption{Evaluation of conflict detectors on both knowledge editing and RAG benchmarks.}
\vspace{-3mm}
\label{tab:detector}
\end{table}

The performance of different conflict detectors is shown in Table~\ref{tab:detector}. The detectors are trained on the training set of MQuAKE for its large scale, and we provide the evaluation results on the test set. As we can see, the detector models can generalize well in the knowledge conflict scenario in RAG datasets, since the task of knowledge editing is closely related to the knowledge conflict scenario. 

\begin{table}[htbp]
\centering
\resizebox{\linewidth}{!}{
\begin{tabular}{lcccc}
\toprule
\textbf{Probe Type} & \textbf{Train Data} & \textbf{$Acc_P$} & \textbf{$F1$} & \textbf{Improve} \\
\midrule
No Probe & N/A & N/A & 61.7 & 0 \\
Random Probe & N/A & 49.6 & 60.3 & -1.4 \\
Weak Probe (2-Layers) & 200 & 56.4 & 65.7 & 4.0 \\
Standard Probe (3-Layers) & 1000 & 86.7 & 74.4 & 12.7 \\
Strong Probe (5-Layers) & 5000 & 87.4 & 76.3 & 14.6 \\
\bottomrule
\end{tabular}
}
\caption{Relationship Between Latent Conflict Probing Accuracy and Final Model Response Quality.}
\vspace{-3mm}
\label{tab:probing-results}
\end{table}

We also analyze the relationship between the accuracy of conflict detection and the quality of the model's responses. We employ probe models of varying sizes combined with different training data volumes, resulting in three distinct probe models ranging from weak to strong capabilities. As shown in Table~\ref{tab:probing-results}, when the probe successfully learns to identify latent conflicts (higher $Acc_P$), the main model achieves a higher $F1$ score, validating that our probing mechanism effectively guides the model toward safer, higher-quality generations. The experimental results indicate that the accuracy of detecting conflicting knowledge within context is positively correlated with the model's ultimate ability to generate faithful responses.

\subsection{Impact of Attention Guidance Loss Item}
\label{sec:attn_analysis}

To further investigate the effect of the hyperparameter $\alpha$, we conduct experiments with multiple values of $\alpha$ and analyze both the attention weights assigned to conflicting knowledge and the corresponding model performance. As shown in Figure~\ref{fig:atten}, increasing $\alpha$ consistently raises the model’s attention, with the growth curve gradually flattening and stabilizing around 0.5. However, model performance does not follow the same trend. Instead, performance peaks when $\alpha$ is in the range of 0.1 to 0.3, after which it declines as $\alpha$ continues to increase. This observation indicates that higher attention to conflicting knowledge does not necessarily lead to better performance. While attending to conflicting knowledge is crucial, the model must also balance its focus on the question itself and other relevant contextual information.

\section{Related Work}
\label{sec:related_work}
Due to space limitations, we provide only a concise overview of the related work here. More detailed discussion can be found in Appendix~\ref{appendix:related_work}.

\textbf{Retrieval-Augmented Generation} (RAG) has emerged as a prominent paradigm for enhancing the factual accuracy and temporal relevance of Large Language Models (LLMs) by incorporating external knowledge sources \citep{shi2024replug, xiang2025use,chen2025you,xiao2025lag, hui2025interact, chen2025mirage}. Even before the emergence of LLMs, researchers were already using RAG technology to improve various natural language processing tasks, such as machine translation~\citep{jiang2022towards, cao2023bridging, gao2024efficient}. Early works such as REALM \citep{guu2020realm} and RAG \citep{lewis2020rag} retrieve relevant passages from large corpora to assist generation. Subsequent research has explored improvements in both the retriever and generator modules, including dense retrieval techniques~\citep{karpukhin2020dense, izacard2022few}, adaptive retrieval strategies~\citep{sun2023recitation, chen2025rethinking}, and hybrid models combining retrieval with parametric memory~\citep{shi2023replug, cao2024retaining, wang2025continuously}. 

\textbf{Contextual Faithfulness} refers to the alignment between the generated output and the provided context, which is especially critical in RAG settings~\citep{huang2025parammute, bi2025parameters, tang2025ssfo}. Prompt-based methods design templates or self-reflection mechanisms to encourage faithful use of context~\citep{selfrag, intuitive}. Decoding-based methods modify generation strategies to enhance the influence of the retrieved context~\citep{coiecd, cad, santosh2025cocolex}. Reinforcement learning frameworks such as CANOE~\citep{si2025teaching} and Context-DPO~\citep{bi2024context} employ an end-to-end paradigm to optimize the generation process and reward contextual faithful response.

\section{Conclusion}
In this work, we focus on how LLMs internally integrate external knowledge with their parametric memory under knowledge conflicts. Through probing-based analysis, we uncovered two insights: conflicting and aligned knowledge states are linearly separable in the model’s latent space, and contextual noise systematically increases the entropy of these representations. Based on this, we introduced a framework named ProbeRAG to improve RAG faithfulness, which combines fine-grained knowledge pruning, latent conflict probing, and conflict-aware attention to enhance contextual faithfulness. Comprehensive experiments across multiple benchmarks and LLMs demonstrate that ProbeRAG consistently outperforms strong baselines. Our framework highlights the importance of explicitly mitigating knowledge conflicts, offering a principled direction for future research.

\section*{Limitations}
While ProbeRAG demonstrates great improvements in textual RAG scenarios, its applicability to multimodal RAG systems remains limited. The current framework is designed around sentence-level textual decomposition and hidden-state probing, which are not directly transferable to modalities such as images, audio, or structured data. In multimodal contexts, knowledge conflicts may manifest in non-textual representations, requiring new strategies for knowledge decomposition, conflict detection, and attention guidance. Extending ProbeRAG to handle heterogeneous modalities would thus require substantial redesign of its probing mechanism and fine-tuning objectives, which we leave as an important direction for future research.

\section*{Acknowledgements}
The project was supported by National Key R\&D Program of China (No. 2022ZD0160501), Natural Science Foundation of Fujian Province of China (No. 2024J011001), and the Public Technology Service Platform Project of Xiamen (No.3502Z20231043). We also thank the reviewers for their insightful comments.

\bibliography{custom}

\newpage
\appendix

\label{sec:appendix}
\section{Frequently Asked Questions (FAQs)}

We summarized some frequently asked questions:

\noindent\textbf{(i) What are the computational costs compared to baselines?}

\begin{figure}
    \centering
    \includegraphics[width=\linewidth]{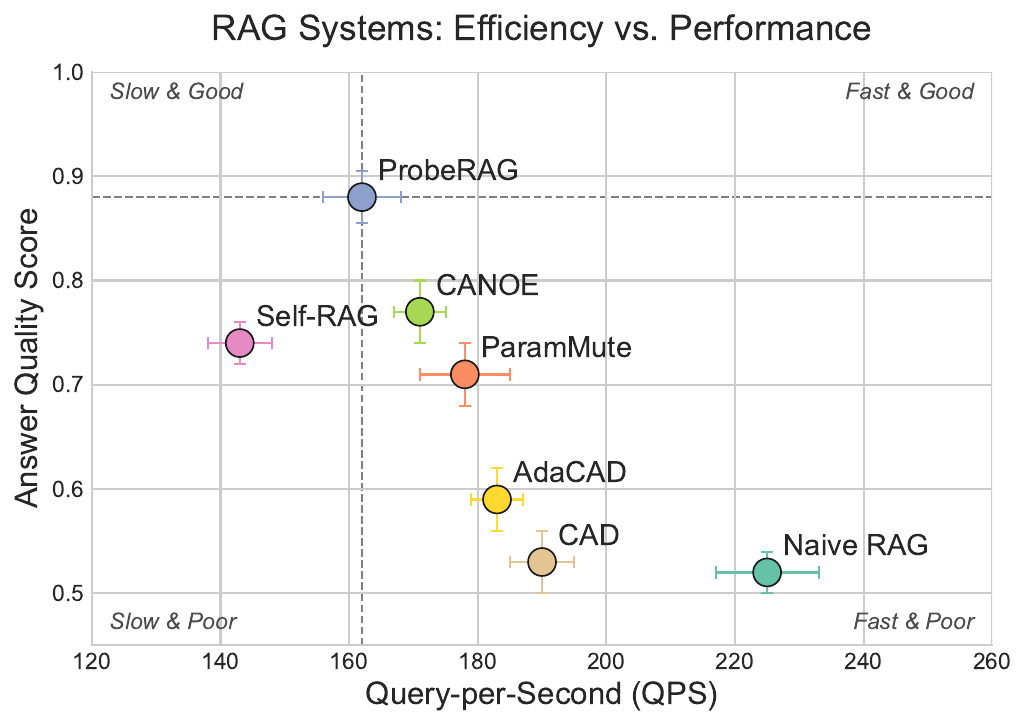}
    \caption{The performance and efficiency comparison of different RAG frameworks on faithfulness.}
    \label{fig:compare}
\end{figure}

As shown in Figure~\ref{fig:compare}, ProbeRAG offers substantially stronger answer quality but does so at the expense of computational efficiency. This trade-off primarily stems from the additional processing steps required by ProbeRAG beyond those in a standard naïve RAG pipeline. As discussed in Section~\ref{sec:method}, ProbeRAG introduces several extra forward passes to explicitly model and analyze the knowledge contained in the retrieved context before producing the final answer.

First, ProbeRAG must encode the retrieved context to extract and decompose fine-grained knowledge statements, which serve as the foundation for downstream conflict analysis. This step requires a full forward pass over the entire context to obtain both the semantic representations and the decomposed knowledge units.

Second, these knowledge statements are jointly fed back into the model in parallel, typically by batching multiple statements into a compressed input sequence. The model then performs another forward pass to produce the corresponding hidden-state representations, which allow ProbeRAG to probe the latent behavior of the model and detect potential conflicts or inconsistencies among pieces of knowledge.

Finally, ProbeRAG integrates these probing results by annotating the original context with the detected conflicting knowledge signals, and this augmented context is passed through the model once more to generate a more faithful and conflict-aware answer. This final forward pass not only incorporates the retrieved evidence but also enables the model to explicitly adjust its reasoning when conflicting information is present.

Although these additional steps introduce a noticeable increase in the computational cost, particularly due to multiple sequential or parallel forward passes, the resulting improvements in reliability and answer quality highlight the effectiveness of ProbeRAG’s design for faithful knowledge integration. In particular, ProbeRAG significantly improves answer faithfulness by explicitly modeling how the LLM internally processes the contextual knowledge, rather than external intervention.

\noindent\textbf{(ii) Will the fine-tuning process lead to less context reliance?}

Existing works~\citep{goyal2024context} have demonstrated that fine-tuning on non-contextual critical data points substantially increases a model’s reliance on its parametric knowledge. Specifically, when a large proportion of training examples are labeled with answers that the model already knows, the fine-tuning gradients disproportionately reinforce these pre-existing parametric priors. As a consequence, the model becomes increasingly confident in retrieving information from its internal memory rather than attending to the external evidence provided in the input. This overreliance on parametric knowledge can be harmful in retrieval-augmented settings: instead of utilizing retrieved context to verify or update its beliefs, the model tends to ignore contextual cues and default to what it has previously memorized, thereby amplifying hallucination risks.

To avoid this failure mode, we train our model on ConFiQA, a dataset primarily composed of counterfactual instances that directly contradict the model’s internal knowledge. Such data points force the model to deviate from its memorized priors and instead rely on the information explicitly given in the context. By repeatedly exposing the model to scenarios where its parametric knowledge is incorrect, we encourage it to strengthen context-grounding behaviors—i.e., to attend to and trust retrieved evidence rather than relying solely on intrinsic memory. As demonstrated in \citet{goyal2024context}, counterfactual data augmentation can meaningfully shift the model’s reliance from parameters to context. Building on this insight, we integrate ConFiQA into the training pipeline as a targeted strategy to mitigate parametric bias and enhance context sensitivity, ultimately improving faithfulness in retrieval-augmented generation.

\noindent\textbf{(iii) How does this work differ from other analyses of knowledge conflicts?}

Some recent studies have observed the some model's attributes, like logits or activation patterns, to analyze the impact of knowledge conflict on RAG faithfulness~\citep{huang2025parammute, zeng2025towards}. However, these attributes are still some other forms of the model's output, which also stand for the model's behavior instead of what the model internally ``think''. Instead, we further deep into the model's latent space, analyzing the internal hidden-state of the model, which represents the fusion of external knowledge processed through the attention mechanism with the model's internal knowledge. The hidden states can truly reflect the internal reasoning processes within the model, representing how the model internally ``thinks'' rather than what the model actually ``does''.

Furthermore, existing methods also differ significantly from ours in the subsequent processing. Some of them change the contextual knowledge~\citep{zeng2025towards}, while others change the model's inherent knowledge~\citep{huang2025parammute}. For contextual noise, filtering is necessary. However, when it comes to conflicting knowledge, we cannot simply modify the model's knowledge, as this would lead to catastrophic forgetting of the model's inherent knowledge, which is the most challenging problem faced in the field of knowledge editing. In this work, we choose to ``guide'' the model instead of ``changing'' the model, by teaching it to pay more attention on the context, especially on the conflicting knowledge, enhancing the model's contextual faithfulness without compromising the its inherent knowledge.

\begin{figure*}
    \centering
    \includegraphics[width=0.8\linewidth]{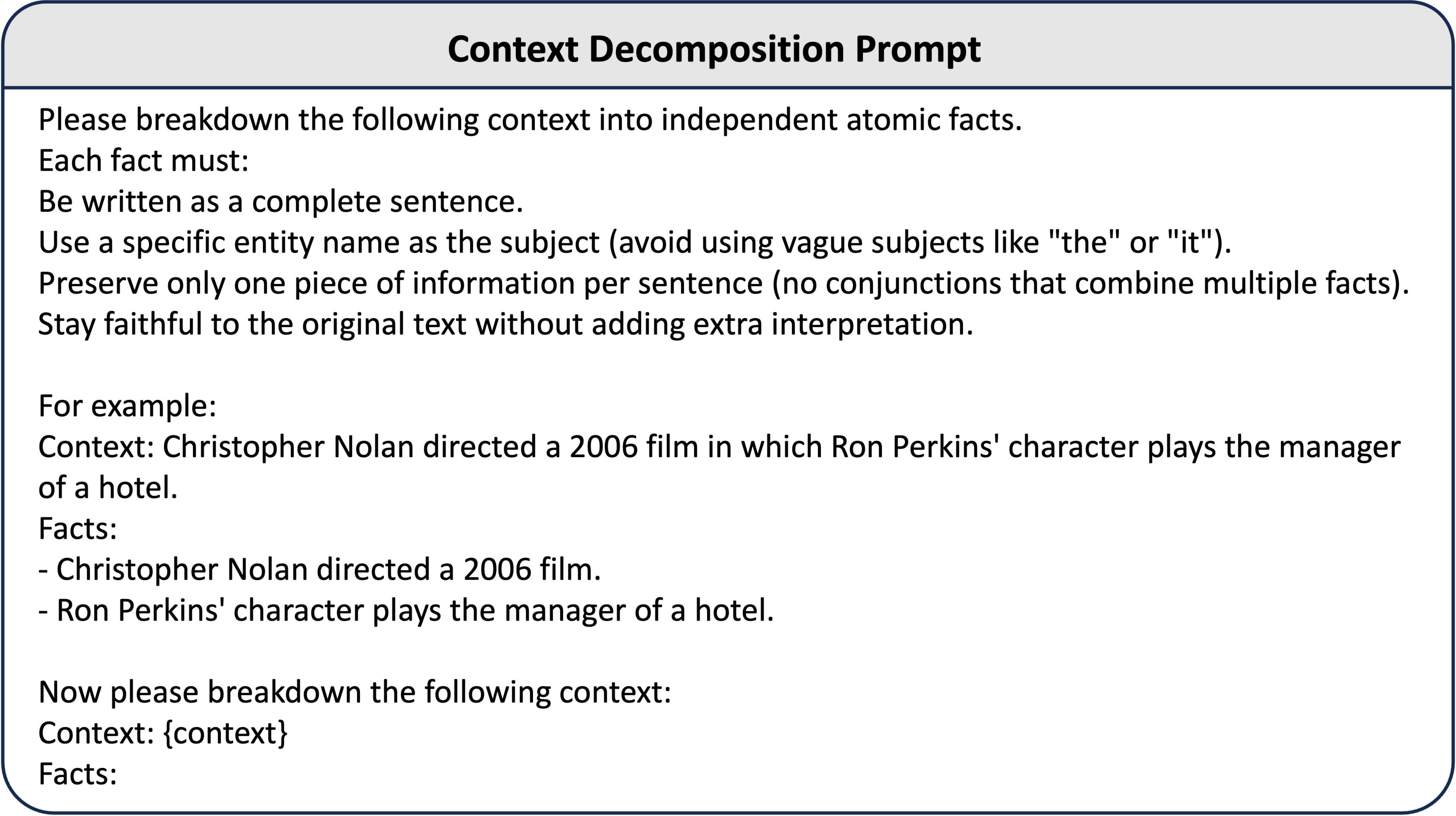}
    \caption{Context decomposition prompt used in the Fine-Grained Knowledge Pruning module.}
    \label{fig:prompt}
\end{figure*}

\section{Algorithmic Description}

The following presents the algorithmic description of the ProbeRAG framework. First, the retrieved context is decomposed into fine-grained knowledge, from which the most relevant ones are selected based on query–knowledge similarity. Second, a hidden-state probe detects conflicts between the selected knowledge and the model’s internal knowledge, and conflicting knowledge is explicitly annotated with special tokens. Third, we introduce conflict-aware attention, which guides the model’s attention on the annotated conflict tokens by incorporating an auxiliary attention-guidance loss into the training objective. The fine-tuned model then generates the final answer conditioned on the pruned and annotated context, enabling more faithful response generation.

\begin{algorithm}
  \SetKwInOut{Input}{Input}
  \SetKwInOut{Output}{Output}
  
  \Input{Question $Q$, retrieved context $D$, model $\mathcal{M}$, probe $\mathcal{P}_\mathcal{M}$}
  \Output{Answer $A$}
  \BlankLine 

  Step 1: Fine-grained Knowledge Pruning\;
  $\{k_1, k_2, ..., k_n\} \leftarrow \text{Decompose}(D)$\;
  \For{$k_i \in \{k_1, k_2, ..., k_n\}$}{
    $s_i \leftarrow \text{similarity}(Q, k_i)$\;
  }
  $S \leftarrow \text{top-k}(\{s_1, s_2, ..., s_n\})$\;
  $D' \leftarrow \{k_i|s_i\in S\}$\;
  
  \BlankLine

  Step 2: Latent Conflict Probing\;
  \For{$k_i \in D'$}{
    $h_i \leftarrow \mathcal{M}(k_i)$\;
    $p_i \leftarrow \mathcal{P}_\mathcal{M}(h_i)$\;
    \If{$p_i > 0.5$}{
        $k_i \leftarrow \text{mark}(k_i)$\;
    }
  }
  \BlankLine

  Step 3: Conflict-Aware Attention\;
  $\mathcal{M}'\leftarrow \text{SFT}(\mathcal{M})$\;
  $A \leftarrow \mathcal{M}'(Q, D')$\;
  \Return{$A$}
  
  \caption{ Workflow of ProbeRAG.}
\end{algorithm}

\section{Implementation Details}
\label{appendix:details}

\paragraph{Detail of ProbeRAG.} For the implementation of ProbeRAG, we configure the experimental settings as follows.
In the Fine-Grained Knowledge Pruning module, we employ GPT-4o to decompose the retrieved context into fine-grained knowledge using the prompt template illustrated in Figure~\ref{fig:prompt}. We then compute semantic similarity among the decomposed knowledge with all-MiniLM-L6-v2 and retain the top-10 most relevant knowledge item.

In the Latent Conflict Probing module, the selected knowledge items are fed into the model, from which we extract hidden states of the decoder. These representations are passed to a trained MLP-based probe for binary classification. The probe consists of three fully connected layers with ReLU activation, followed by a sigmoid normalization. For training the probe, we sample 1,000 instances with a learning rate of 0.001 and run for 10 epochs.

For the Conflict-Aware Attention module, we set the weighting hyperparameter $\lambda = 0.1$. On the ConFiQA dataset, we allocate 13,500 instances for training (with 4,500 samples each from the MC, MR, and QA subsets), while the remaining data are reserved for evaluation. We fine-tune the model using LoRA, where the rank $r$ is set to 16, the scaling factor $\alpha$ to 16, and the learning rate to $3\times 10^{-5}$, training for a total of 5 epochs. Finally, during inference, we set the temperature parameter to 0 to ensure reproducibility of results. Detailed hyperparameters are listed in Table~\ref{tab:hyperparameters}.

\begin{table}[ht]
\centering
\resizebox{\linewidth}{!}{
\begin{tabular}{@{}llll@{}}
\toprule
\textbf{Category} & \textbf{Hyperparameter} & \textbf{Symbol} & \textbf{Value} \\ \midrule
\multirow{6}{*}{\textbf{Training}} & Learning Rate & $\eta$ & $3 \times 10^{-5}$ \\
 & Batch Size & $B$ & 4 \\
 & Epochs & $N$ & 3 \\
 & Weight Decay & - & 0.01 \\
 & Optimizer & - & AdamW \\
 & Warmup Ratio & - & 0.1 \\ \midrule
\textbf{Loss Function} & \textbf{Attention Loss Weight} & $\mathbf{\lambda}$ & \textbf{0.1} \\ \midrule
\multirow{4}{*}{\textbf{Attention}} & \textbf{Aggregated Layers} & - & \textbf{Last 1 Layer ([-1])} \\
 & \textbf{Aggregated Heads} & - & \textbf{All} \\
 & Threshold & $\epsilon$ & $1 \times 10^{-8}$ \\ \bottomrule
\end{tabular}
}
\caption{Hyperparameters for conflict-aware attention.}
\label{tab:hyperparameters}
\end{table}

\paragraph{Detail of Baseline Implementations.} For all baselines reported in the main experiments, we adopt a sampling temperature of 0 and a maximum generation length of 128 tokens. For CAD, we set the hyperparameter $\alpha$ = 0.9. For all prompt-based methods, we directly employ the prompt templates provided in the original papers. For all training-based methods, we use the same training data as ProbeRAG, sampled from ConFiQA. Specifically, for Context-DPO, we apply the same LoRA configuration during training. For CANOE, we follow the original training setup and perform full-parameter fine-tuning on 4 NVIDIA A100 GPUs.

\paragraph{Detail of Ablation Study.} 

For the w/o Knowledge Pruning variant, we partition the input context directly into sentences and subsequently apply the conflict detection module to determine whether each sentence conflicts with the model’s parametric knowledge. For the w/o Conflict Detection variant, we fine-tune the model using the decomposed knowledge directly. Since conflicting knowledge is not explicitly identified, only the loss term $\mathcal{L}_{\text{LM}}$ is active during Conflict-Aware Attention. For the w/o Fine-Tuning variant, we remove the $\mathcal{L}_{\text{Attn}}$ term, which reduces the training objective to standard SFT without attention-level supervision.

\begin{figure*}[htbp]
  \centering
    \includegraphics[width=0.8\linewidth]{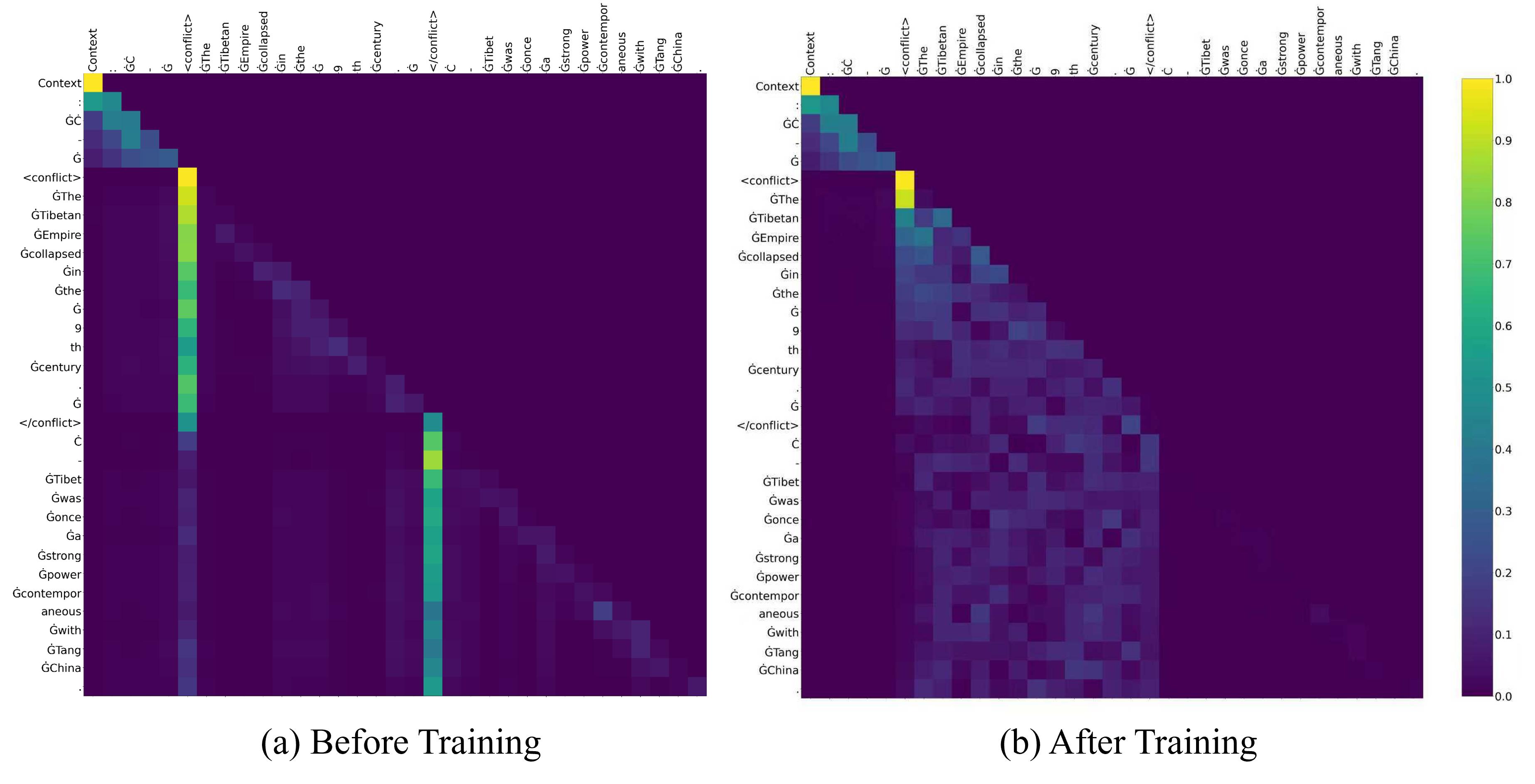}
    \caption{Heatmap visualization of attention weight distribution.}
    \label{fig:heatmap}
\end{figure*}

\section{Additional Experiments}

\subsection{Probing Accuracy and Final Performance}

In this subsection, we analyze the relationship between the accuracy of conflict detection and the quality of the model's responses. To control the accuracy of probe detection, we employ probe models of varying sizes combined with different training data volumes, resulting in three distinct probe models ranging from weak to strong capabilities. As shown in Table~\ref{tab:probing-results}, when the probe successfully learns to identify latent conflicts (higher $Acc_P$), the main model achieves a higher $F1$ score, validating that our probing mechanism effectively guides the model toward safer, higher-quality generations. Notably, using a ``Random Probe'' results in performance similar to ``No Probe'', reinforcing that accurate diagnostic feedback is essential. The experimental results indicate that the accuracy of detecting conflicting knowledge within context is positively correlated with the model's ultimate ability to generate faithful responses. Therefore, the performance of the probe model is central to the ProbeRAG framework.

\subsection{Attention Heatmap Visualization}

To visually validate the effectiveness of our proposed framework, we conducted a visualization analysis of the attention weights before and after model training. The experiment selected the LLaMA-3.1-8B-Instruct model as the subject of study and fed it context containing conflict knowledge with special annotations. Specifically, we use the $\langle conflict \rangle$ and $\langle /conflict \rangle$ tags to explicitly mark conflict information within the text. 

We visualized the attention heatmap in Figure~\ref{fig:heatmap} and annotated the specific numerical ranges of attention weights corresponding to tokens of different categories in Table~\ref{tab:attention_weight}. As shown in Figure~\ref{fig:heatmap}, we observe significant differences between the two stages. Before training, the model's attention distribution shows no particular focus on content within the $\langle conflict \rangle$ tag. Its attention patterns primarily followed linguistic habits formed during the pre-training phase. However, after fine-tuning with our proposed ``attention-guided loss'' function, the model exhibits significantly enhanced attention weights on tokens enclosed by the $\langle conflict \rangle$ tag when processing the same input. The highlighted regions in the heatmap clearly demonstrate that the model has learned to allocate more computational resources to these explicitly labeled conflict knowledge fragments.

\begin{table}[htbp]
\centering
\resizebox{\linewidth}{!}{
\begin{tabular}{lll}
\toprule
\textbf{Token Category} & \textbf{Before Training} & \textbf{After Training} \\ 
\midrule
\textbf{Conflict Context} & $\sim$0.14 - 0.23 & \textbf{$\sim$0.31 - 0.61} \\ 
\midrule
\textbf{Special Tokens} & \textbf{$\sim$0.61 - 0.85} & $\sim$0.23 - 0.46 \\ 
\midrule
\textbf{General Context} (Other) & $\sim$0.15 - 0.21 & $\sim$0.17 - 0.28 \\ 
\bottomrule
\end{tabular}
}
\caption{Attention Weight Distribution for Different Token Categories: Before vs. After Training}
\label{tab:attention_weight}
\end{table}

This experimental result provides compelling evidence for the core design intent of our framework. The phenomenon of attention weights focusing is not coincidental, but rather a direct manifestation of our attention-guided loss function successfully influencing the model's internal mechanisms. By imposing penalties, this loss function guides the model to proactively identify and prioritize processing these critical conflicting information during training. Consequently, this visualization analysis offers robust qualitative evidence for our approach, confirming its effectiveness in directing the model's attention toward specific knowledge domains.

\subsection{Analysis on Additional Models}

\begin{table*}[t]
\centering
\small
\resizebox{0.9\textwidth}{!}{
\begin{tabular}{lcccccccccc}
\toprule
 \multirow{2}{*}{\textbf{Method}} & \multicolumn{2}{c}{\textbf{FaithEval}} & \multicolumn{2}{c}{\textbf{ConFiQA (MC)}} & \multicolumn{2}{c}{\textbf{ConFiQA (MR)}} & \multicolumn{2}{c}{\textbf{ConFiQA (QA)}} & \multicolumn{2}{c}{\textbf{SQuAD}} \\
\cmidrule(lr){2-3} \cmidrule(lr){4-5} \cmidrule(lr){6-7} \cmidrule(lr){8-9} \cmidrule(lr){10-11}
 & F1 & EM & F1 & EM & F1 & EM & F1 & EM & F1 & EM \\
\midrule
\multicolumn{11}{c}{\cellcolor{orange!10}\textbf{LLaMA-2-7B-Chat-HF}} \\

 Context-DPO & 63.2 & 50.7 & 57.9 & 32.0 & 58.5 & 32.7 & 73.7 & 64.7 & 62.4 & 41.8 \\
 CANOE & \textbf{70.6} & 52.3 & 73.9 & \textbf{70.2} & 75.2 & 72.6 & 74.3 & 72.7 & 63.2 & 45.6 \\
 ProbeRAG & 68.3 & \textbf{54.4} & \textbf{79.1} & 69.7 & \textbf{80.2} & \textbf{77.0} & \textbf{86.1} & \textbf{81.7} & \textbf{65.4} & \textbf{52.1} \\
\midrule
\multicolumn{11}{c}{\cellcolor{orange!10}\textbf{Qwen2.5-7B-Instruct}} \\
 Context-DPO & 65.1 & 50.2 & 62.7 & 53.7 & 71.1 & 58.8 & 75.0 & 66.3 & 55.2 & 36.4 \\
 CANOE & \textbf{68.1} & \textbf{53.9} & 68.7 & 61.1 & 71.7 & 67.8 & 70.6 & 66.9 & 59.4 & 41.3 \\
 ProbeRAG & 63.5 & 48.9 & \textbf{88.8} & \textbf{86.2} & \textbf{89.6} & \textbf{86.2} & \textbf{94.3} & \textbf{91.5} & \textbf{61.6} & \textbf{46.2} \\
\bottomrule
\end{tabular}
}
\caption{Supplementary experimental results on additional model architectures.}
\label{tab:more_model}
\end{table*}

Table~\ref{tab:more_model} presents supplementary results on two additional model architectures, LLaMA-2-7B-Chat-HF and Qwen2.5-7B-Instruct, evaluated across multiple benchmarks. Consistent with the main findings, ProbeRAG demonstrates notable improvements over both Context-DPO and CANOE, particularly on conflict-sensitive datasets such as ConFiQA and FaithEval. For LLaMA-2-7B-Chat-HF, ProbeRAG achieves the highest scores on most ConFiQA variants, while also maintaining competitive performance on FaithEval and SQuAD.

On Qwen2.5-7B-Instruct, the advantage of ProbeRAG becomes even more pronounced: it consistently outperforms both baselines across all ConFiQA settings, with substantial gains in F1 and EM. Although CANOE occasionally remains competitive on less conflict-intensive benchmarks, ProbeRAG shows strong generalization in resolving conflicting knowledge. These results confirm that the effectiveness of ProbeRAG extends beyond a single backbone, underscoring its robustness across different instruction-tuned LLMs.

\subsection{Impact on Conflict Markers}

\begin{table}[htbp]
\centering
\resizebox{\linewidth}{!}{
\begin{tabular}{llcc}
\toprule
\textbf{Method} & \textbf{Context Condition} & \textbf{F1} & \textbf{Exact Match} \\ 
\midrule
\textbf{Vanilla LLM}  & No Markers                 & 45.2 & 38.1 \\ 
\midrule
\textbf{Prompt-based} & With Markers               & 62.5 & 55.4 \\ 
\midrule
\textbf{ProbeRAG}     & \textbf{With Markers (Original)} & \textbf{88.3} & \textbf{85.9} \\ 
\midrule
\textbf{ProbeRAG}     & \textbf{No Markers (Ablation)}   & 74.3 & 71.4 \\ 
\bottomrule
\end{tabular}
}
\caption{Ablation Study on Conflict Marker.}
\label{tab:ablation_marker}
\end{table}

In the ProbeRAG framework, we introduced special tokens to mark conflicting knowledge within the context. In this subsection, we delve into the impact of introducing these special tokens. Specifically, we conducted ablation experiments to compare performance before and after their introduction. The experimental results are shown in Table~\ref{tab:ablation_marker}. With conflict markers indicating conflicting knowledge in the context, LLMs can focus more on this aspect, thereby enhancing the faithfulness of their responses. Notably, with respect to the ProbeRAG framework, conflict markers can further guide attention distribution and assist in calculating loss terms to guide subsequent training, leading to further performance improvements.

\subsection{Supplementary Experimental Results}

Table~\ref{tab:alpha_results} reports the detailed numerical results corresponding to Figure~\ref{fig:atten}, including both the model accuracy and the attention weight assigned to conflicting knowledge across different values of $\alpha$ for LLaMA-3.1-8B-Instruct, Qwen3-8B, and Mistral-7B-v0.3. Consistent with the trends shown in the figure, attention weights increase steadily with larger $\alpha$, saturating around $\alpha=0.5$. In contrast, accuracy peaks within a smaller range of $\alpha$ (0.1--0.3) and then declines as $\alpha$ continues to grow. These results highlight that while higher $\alpha$ values encourage stronger focus on conflicting knowledge, this emphasis can come at the cost of overall performance. The tabulated results thus provide a more fine-grained view of the trade-off between model attention allocation and accuracy.

\begin{table}[t]
  \centering
  \resizebox{1\linewidth}{!}{
  \begin{tabular}{c|cc|cc|cc}
    \toprule
    \multirow{2}{*}{$\alpha$} & 
    \multicolumn{2}{c|}{LLaMA-3.1-8B-Instruct} & 
    \multicolumn{2}{c|}{Qwen3-8B} & 
    \multicolumn{2}{c}{Mistral-7B-v0.3} \\
    \cmidrule(lr){2-3} \cmidrule(lr){4-5} \cmidrule(lr){6-7}
     & Acc & Attention & Acc & Attention & Acc & Attention \\
    \midrule
    0.0 & 0.552 & 0.020 & 0.512 & 0.115 & 0.631 & 0.070 \\
    0.1 & 0.644 & 0.105 & 0.604 & 0.022 & 0.663 & 0.193 \\
    0.2 & 0.632 & 0.188 & 0.573 & 0.195 & 0.598 & 0.203 \\
    0.3 & 0.635 & 0.231 & 0.639 & 0.283 & 0.559 & 0.211 \\
    0.4 & 0.554 & 0.331 & 0.495 & 0.415 & 0.611 & 0.314 \\
    0.5 & 0.482 & 0.442 & 0.443 & 0.390 & 0.538 & 0.463 \\
    0.6 & 0.333 & 0.464 & 0.430 & 0.381 & 0.289 & 0.543 \\
    0.7 & 0.201 & 0.483 & 0.153 & 0.474 & 0.171 & 0.549 \\
    0.8 & 0.214 & 0.481 & 0.211 & 0.444 & 0.117 & 0.459 \\
    0.9 & 0.210 & 0.464 & 0.207 & 0.457 & 0.194 & 0.538 \\
    \bottomrule
  \end{tabular}
  }
  \caption{Accuracy and Attention Weight across different $\alpha$ values for three models.}
  \label{tab:alpha_results}
\end{table}

\section{Case Study}

In this section, we present a case study to further illustrate how our proposed framework ProbeRAG enforces contextual faithfulness under knowledge conflicts. We conduct the analysis on the Faitheval dataset using the LLaMA-3.1-8B-Instruct model, and the results are shown in Table~\ref{tab:case}. ProbeRAG first decomposes the retrieved context into fine-grained knowledge, followed by filtering and conflict detection. As indicated in the table, the context explicitly states that construction speed is the dominant benefit of seismic testing, whereas the model’s prior knowledge typically associates seismic testing with structural safety. Through our conflict detection probe, ProbeRAG successfully identifies such conflicts and, with the aid of Conflict-Aware Attention, reinforces the model’s attention to the conflicting knowledge (3) and (5). As a result, ProbeRAG generates the correct answer, \emph{“Buildings will be built faster,”} which faithfully reflects the contextual evidence rather than relying on the model’s internal knowledge. This case study highlights the effectiveness of our framework in ensuring contextual faithfulness in scenarios involving contextual knowledge conflicts.

\begin{table*}[t]
\centering
\small

\resizebox{\linewidth}{!}{
\begin{tabular}{p{2cm} p{12cm}}
\toprule
\textbf{Question} & A group of engineers wanted to know how different building designs would respond during an earthquake. They made several models of buildings and tested each for its ability to withstand earthquake conditions. Which will most likely result from testing different building designs? \\
\midrule
\textbf{Context} & Seismic testing of building models is crucial for understanding how structures will behave during earthquakes. Engineers approach these tests with a myriad of designs, each aiming to improve certain aspects of building performance, such as safety, aesthetic appeal, and construction speed... \\
\midrule
\textbf{Knowledge Extracted} & 
(1) Seismic testing of building models is crucial for understanding structural behavior during earthquakes. \newline
(2) Engineers approach tests with a myriad of designs aiming to improve safety, aesthetic appeal, and construction speed. \newline
(3) $\langle conflict \rangle$ \textbf{Implementation of efficient techniques can enhance building times by up to 30\%.} $\langle/ conflict \rangle$ \newline
(4) Seismic testing aligns efficiency with safety in contemporary civil engineering practices. \newline
(5) $\langle conflict \rangle$ \textbf{Speed of construction is a dominant benefit of testing building designs under earthquake simulation conditions.} $\langle/ conflict \rangle$ \newline
(6) Optimization of construction speed guarantees resilience and rapid realization of new buildings through continued innovation and testing. \newline
...\\
\midrule
\textbf{Model Answer} & Buildings will be built faster. \\
\bottomrule

\end{tabular}
}
\caption{Case Study Result. This table displays the knowledge extracted from the context and the results of identifying knowledge conflicts. Based on the conflicting knowledge annotations, the model trained with conflict-aware attention can correctly answer questions with contextual faithfulness.}
\vspace{-3mm}
\label{tab:case}
\end{table*}

\section{Related Work}
\label{appendix:related_work}

\paragraph{Retrieval-Augmented Generation}
RAG has become a cornerstone paradigm to improve the factual reliability and adaptability of LLMs by explicitly integrating external information during the generation process~\citep{yao2023react, wang2025unveiling, nafar2025learning, sharma2025retrieval}. Early contributions such as REALM~\citep{guu2020realm} and RAG~\citep{lewis2020rag} pioneered the idea of end-to-end frameworks in which a retriever component selects relevant passages from large-scale corpora, which are then consumed by a generator to produce responses grounded in retrieved evidence. This framework demonstrate clear advantages over purely parametric models, particularly in tasks requiring factual precision or recent event knowledge~\citep{math13050856}.

Following these foundational works, the research community has proposed a series of improvements targeting both the retriever and generator components~\citep{chen2025improving}. For retrieval, dense retrieval methods~\citep{karpukhin2020dense, izacard2022few} introduced learned embeddings that outperform traditional sparse methods (e.g., BM25) in capturing semantic relevance. Subsequent refinements incorporated multi-vector representations~\citep{santhanam2021colbertv2}, passage reranking~\citep{nogueira2019passage}, and adaptive retrieval strategies~\citep{sun2023recitation}, where the retrieval budget is dynamically allocated based on the complexity of the query or the uncertainty of the model’s predictions.

On the generator side, researchers have explored how to more effectively incorporate retrieved passages during decoding. FiD (Fusion-in-Decoder)~\citep{izacard2020leveraging} demonstrated the effectiveness of late-fusion mechanisms, where a Transformer decoder attends jointly over multiple retrieved documents. Later works extended this paradigm with hierarchical fusion~\citep{ram2023context}, sparse attention mechanisms~\citep{shuster2022language}, and multi-hop retrieval pipelines~\citep{xu2023retrieval}. Hybrid models such as RePlug~\citep{shi2023replug} and Retro~\citep{borgeaud2022improving} further integrated retrieval into pretraining or finetuning pipelines, blending parametric and non-parametric memories to achieve both scalability and factual accuracy. More recently, adaptive frameworks~\citep{chen2024adaptive} proposed fine-grained controls over how retrieval signals are weighted depending on task type, query ambiguity, or user intent.

In addition to architectural innovations, researchers have also investigated the evaluation and efficiency of RAG systems. Benchmarks such as KILT~\citep{petroni2020kilt} and ELI5~\citep{fan2019eli5} standardize evaluation across knowledge-intensive tasks, while efficiency-focused studies~\citep{guu2020retrieval} highlight the trade-off between accuracy, latency, and resource consumption.

\paragraph{Contextual Faithfulness}
Contextual faithfulness, defined as the degree to which model outputs remain consistent with retrieved or provided context~\citep{niu2024ragtruth, wang2025retrieval}, has emerged as a central concern in RAG research~\citep{chern2023factool, li2024attributionbench}. Without explicit mechanisms to enforce faithfulness, models may hallucinate, overgeneralize, or generate outputs inconsistent with retrieved passages~\citep{chen2024knowledge, sui2025fidelis}.

Prompt-based methods were among the earliest to address this challenge~\citep{choi2025conflict}. Self-RAG~\citep{selfrag} introduced self-reflection mechanisms, where models generate justifications for retrieved content and use these to re-ground their outputs. Template-based prompting approaches~\citep{intuitive} designed structured query-response formats to encourage explicit grounding, though such methods often struggle with generalization across tasks.

Decoding-based approaches tackle faithfulness by modifying the generation process itself~\citep{wang2025adacad, khandelwal2025cocoa}. Contrastive Decoding~\citep{coiecd} and Context-Aware Decoding (CAD)~\citep{cad} explicitly re-weight token probabilities during beam search to favor outputs aligned with retrieved context. Similarly, likelihood re-ranking techniques~\citep{zhang2024retrievalqa} compare candidate responses against retrieved evidence to penalize hallucinations. These approaches maintain the flexibility of generation while reducing unfaithful responses.

Reinforcement learning (RL) has also been extensively applied to enhance contextual faithfulness~\citep{huang2025improving, li2025generate, duong2025scope, bai2025understanding, huangyw2025dynamic}. CANOE~\citep{si2025teaching} integrates reward models that explicitly score the grounding of responses in retrieved passages. Context-DPO~\citep{bi2024context} extends direct preference optimization to context-aware settings, allowing LLMs to directly learn from pairwise comparisons of faithful versus unfaithful outputs. Such RL-based frameworks emphasize end-to-end optimization, reducing reliance on handcrafted prompts or decoding heuristics.

Beyond methodological innovations, recent surveys~\citep{zhou2023context, ji2023survey, xiang2026agentic} highlight persistent challenges in faithfulness evaluation. Automatic metrics such as factual consistency~\citep{thorne2018fever} or entailment-based scores~\citep{falke2019ranking, guo2023evaluating} provide useful proxies but often fail to capture nuanced inconsistencies or omissions. Consequently, many works advocate for human-in-the-loop evaluation frameworks to assess contextual grounding at scale.

\paragraph{Knowledge Conflict}
Knowledge conflict arises when the retrieved evidence contradicts either the model’s internal memory or other retrieved documents, creating ambiguity in determining which to trust~\citep{jin2024tug}. This problem is particularly acute in dynamic knowledge environments, where information evolves over time or when sources exhibit factual inconsistency~\citep{manakul2023self}.

A growing body of work has investigated mechanisms to detect, represent, and resolve knowledge conflicts. Astute RAG~\citep{wang2025astute} introduces a source-aware retrieval module, leveraging reliability estimation to assess which sources are more trustworthy in the face of contradictions. FaithfulRAG~\citep{zhang2025faithfulrag} explicitly models fact-level conflicts, decomposing retrieved evidence into atomic claims and guiding the generation process through a self-thinking phase that resolves inconsistencies.

Alternative approaches focus on information-theoretic principles. Swin-VIB~\citep{wang2025accommodate}, for example, applies a variational information bottleneck to modulate the trade-off between fidelity to retrieved evidence and reliance on internal knowledge, thereby accommodating conflicts in a principled manner. Other works~\citep{xu2024knowledge} categorize conflicts into types, such as temporal drift, factual contradiction, or perspective variance, and tailor resolution strategies accordingly.

Recent research also extends conflict resolution beyond the text domain. Multimodal RAG systems~\citep{gao2023examining, xu2024survey, yang2026graph} face analogous challenges, as retrieved visual or audio evidence may not align with textual outputs. This motivates broader frameworks for consistency checking across modalities. Furthermore, evaluation efforts~\citep{xu2024knowledge} emphasize the need for standardized benchmarks that explicitly include conflict scenarios, enabling more systematic analysis of models’ conflict-handling behaviors.

In summary, while significant progress has been made, knowledge conflict remains an open problem. Robust handling of contradictory information is critical not only for improving factual accuracy but also for building user trust in RAG-based systems deployed in real-world applications.

\section{The Use of Large Language Models}

In preparing this paper, we made limited use of Large Language Models (LLMs). Specifically, LLMs were employed for two purposes: (i) to aid in polishing the writing by improving grammar, readability, and clarity without altering the scientific content, and (ii) to assist in retrieval and discovery tasks, such as identifying and organizing related work. No LLMs were used for generating novel research ideas, designing experiments, or analyzing results. All conceptual and technical contributions presented in this paper are the sole work of the authors.

\section{Reproducibility Statement}
Our code, datasets, and implementation details are anonymously available at \textcolor{blue}{\url{https://github.com/XMUDeepLIT/ProbeRAG}}.
We make significant efforts to ensure the reproducibility of our work. The details of model architectures, hyperparameters, and training settings are provided in Section~\ref{sec:setup} of the main paper. Additional implementation details and full experimental setups are provided in Appendix~\ref{appendix:details}.

\section{Ethics statement}
This work does not involve any experiments with human subjects, sensitive personal data, or information that could identify individuals. All datasets used in our experiments are publicly available and commonly adopted in prior research. We carefully follow dataset licenses and ensure that no proprietary or private information is disclosed. We acknowledge potential concerns regarding bias and fairness in LLMs and retrieval corpora, and we provide detailed dataset descriptions in the appendix to facilitate transparent evaluation.

\end{document}